\documentclass[journal]{IEEEtran}

\usepackage{xcolor,soul,framed} 

\colorlet{shadecolor}{yellow}
\usepackage[pdftex]{graphicx}
\graphicspath{{../pdf/}{../jpeg/}}
\DeclareGraphicsExtensions{.pdf,.jpeg,.png}

\usepackage[cmex10]{amsmath}
\usepackage{array}
\usepackage{mdwmath}
\usepackage{mdwtab}
\usepackage{eqparbox}
\usepackage{url}
\usepackage{diagbox}
\usepackage{graphicx}
\usepackage{microtype}

\usepackage{tipa}
\usepackage{hyperref}
\usepackage{multirow}

\begin{document}
\bstctlcite{IEEEexample:BSTcontrol}
    \title{Cueless EEG Imagined Speech for Subject Identification: Dataset and Benchmarks}

  \author{Ali~Derakhshesh,
        Zahra~Dehghanian,  
        Reza~Ebrahimpour,  and~Hamid~R.~Rabiee,~\IEEEmembership{Senior Member,~IEEE}
\thanks{A. Derakhshesh and Z. Dehghanian are with the Department of Computer Engineering, Sharif University of Technology, Tehran, Iran.}
\thanks{R. Ebrahimpour is with the Center for Cognitive Science, Institute for Convergence Science and Technology (ICST), Sharif University of Technology, Tehran, Iran.}%
\thanks{H. R. Rabiee (corresponding author) is with the Department of Computer Engineering, Sharif University of Technology, Tehran, Iran (e-mail: rabiee@sharif.edu).}%
}

\markboth{IEEE TRANSACTIONS ON BIOMETRICS, BEHAVIOR, AND IDENTITY SCIENCE, VOL.~x, NO.~x, February~2025
}{Derakhshesh \MakeLowercase{\textit{et al.}}: Cueless EEG imagined speech for subject identification: dataset and benchmarks}

\maketitle

\begin{abstract}
Electroencephalogram (EEG) signals have emerged as a promising modality for biometric identification. While previous studies have explored the use of imagined speech with semantically meaningful words for subject identification, most have relied on additional visual or auditory cues to enhance accuracy. In this study, we introduce a cueless EEG-based imagined speech paradigm, where subjects imagine the pronunciation of semantically meaningful words without any external cues related to the target word. This innovative approach addresses the limitations of prior methods by requiring subjects to select and imagine words from a predefined list in a natural manner. The dataset comprises over 4,350 trials from 11 subjects across five sessions. We assess a variety of classification methods, including traditional machine learning techniques such as Support Vector Machines (SVM) and XGBoost, as well as time-series foundation models and deep learning architectures designed explicitly for EEG classification, such as EEG Conformer and Shallow ConvNet. A session-based hold-out validation strategy was employed to ensure reliable evaluation and prevent data leakage. Our results demonstrate outstanding classification accuracy, reaching 99.44\%. These findings highlight the potential of cueless EEG paradigms for secure and reliable subject identification in real-world applications, such as brain-computer interfaces (BCIs).
\end{abstract}


%
\IEEEpeerreviewmaketitle


\section{Introduction}

\IEEEPARstart{B}{iometric}
Security systems could be developed by integrating biometric databases with recognition techniques. This system works by capturing biometric information from individuals, extracting a set of discriminative features, and matching these features against pre-stored templates within the biometric database \cite{moctezuma2019subjects}.
A significant number of researchers employ machine learning or deep learning models as the aforementioned recognition techniques, utilizing datasets composed of biometric samples from individuals.
These systems could be developed using two distinct approaches: subject identification and subject authentication, which Bidgoly et al. \cite{bidgoly2020survey} clearly distinguished. In subject identification, each individual is treated as a separate class, and the model performs multi-class classification to assign the input to one of these classes. On the other hand, in subject authentication, the model functions as a binary classifier to either accept or deny the claimed identity of the input.

In biometric subject recognition literature, various types of biometric data have been utilized. For instance, Grosz et al. \cite{grosz2023afr} used fingerprint data and developed a deep learning architecture named AFR-Net for both subject identification and authentication. Alay et al. \cite{alay2020deep} utilized multimodal biometrics, including iris, face, and finger vein data, and implemented a CNN-based model for subject identification. Additionally, other studies have investigated the use of palmprints \cite{trabelsi2022efficient} and ECG signals \cite{fatimah2022biometric} for subject identification.

Electroencephalogram (EEG) is a bioelectric signal generated by neural processes in the brain, carrying extensive information about the brain activity.
The acquisition methods of EEG are classified into surface EEG and deep EEG. Surface EEG refers to the use of electrodes on the scalp to record the brain's electrical activity \cite{qin2023application}.
In this work, when we refer to EEG, we specifically mean surface EEG.
The signal recorded by EEG has poor spatial resolution and a low signal-to-noise ratio; however, due to its safety and high temporal resolution in capturing neural signal's dynamic changes, it remains popular among researchers \cite{eeg_properties}.

EEG has attracted significant interest as a biometric for subject identification due to its unique advantages. Unlike fingerprints, which can be forged or spoofed, EEG is difficult to replicate. In coercive situations, EEG patterns change, allowing systems to detect if a subject is under duress. Furthermore, EEG ensures the presence of a living subject, as a dead body cannot produce EEG signals \cite{bidgoly2020survey}.

Here, we design a cueless EEG imagined speech paradigm and provide a dataset based on this paradigm, which includes multiple sessions for each subject. We then implement a classification framework to identify subjects from this dataset using both machine learning and deep learning approaches.

These findings make three key contributions: (1) introducing a more naturalistic cueless paradigm for EEG data collection, (2) providing a new high-quality dataset for the research community, and (3) establishing comprehensive benchmarks using current state-of-the-art methods.

\section{Related works}
Researchers have employed various EEG-based paradigms for identifying subjects through brain signals. Some have employed the resting-state EEG paradigm, with eyes-closed and eyes-open conditions, similar to the work of Di et al. \cite{di2019robustness}. Another commonly used paradigm is motor imagery, where Das et al. \cite{das2018motor} identified subjects based on EEG signals recorded during imagined movements of the right arm, left arm, right leg, and left leg. Additionally, other paradigms, such as steady-state visual evoked potentials (SSVEPs) \cite{piciucco2017steady} and eye blinking \cite{abo2015new}, have also been explored for subject identification.

Imagined speech as a speech-related paradigm has also been utilized for subject identification. Nieto et al. \cite{thinking_out_loud} distinguished various speech-related paradigms and defined imagined speech as the motor imagery of speaking, where participants are required to feel as though they are producing speech focusing on the mental simulation of articulatory gestures without any actual physical movements.

The imagined speech paradigm offers several key advantages for biometric identification when compared to the other paradigms previously mentioned. A primary benefit is that imagined speech is an entirely covert and internally generated process. This stands in contrast to event-related potential (ERP) based paradigms, which depend on external stimuli to evoke a neural response. SSVEP systems, for example, require users to gaze at flashing visual stimuli, a method that can induce eye strain and poses a known risk for individuals with photosensitive epilepsy. Furthermore, unlike motor imagery, which is typically limited to a few imagined limb movements, or eye blinking, which offers minimal variation, imagined speech allows for a virtually limitless vocabulary of words or phrases.

Brigham et al. \cite{brigham2010subject} explored imagined speech for subject identification using the syllables /ba/ and /ku/. However, Moctezuma et al. \cite{moctezuma2019subjects} raised concerns about this approach for not incorporating words with semantic meaning and subsequently introduced a new dataset featuring the imagined words "up," "down," "left," "right," and "select" in Spanish. In their study, they employed a random forest classifier on discrete wavelet transform energy coefficients to identify subjects. Despite their contributions, their model evaluation approach has limitations, as they used 10-fold cross-validation, which involves shuffling the data. Since their dataset contains only one session per subject, shuffling results in training and test samples being drawn from the same session failing to demonstrate the model's robustness across sessions over time.

Furthermore, several other datasets containing imagined speech of words with semantic meanings are available, as summarized in Table \ref{tab:imagined_datasets}.

\begin{table*}[h!]
    \centering
    \caption{EEG-based imagined speech datasets featuring words with semantic meanings.}
    \label{tab:imagined_datasets}
    \begin{tabular}{|l|c|c|c|}
    \hline
    \textbf{Dataset} & \textbf{Language} & \textbf{Cue Type} & \textbf{Target Words / Commands} \\ \hline

    Coretto et al. \cite{coretto2017_dataset} & Spanish & Visual + Auditory & 
    up, down, right, left, forward, backward \\ \hline

    Asghari Bejestani et al. \cite{asghari2022_dataset} & Persian & Auditory & 
    up, down, left, right, yes, no \\ \hline

    Qureshi et al. \cite{qureshi2017_dataset} & English & Auditory & 
    go, back, left, right, stop \\ \hline

    BCI Competition 2020 Track3 \cite{bci_competition_dataset} & English & Auditory & 
    hello, help me, stop, yes, thank you \\ \hline

    \end{tabular}
\end{table*}

While these studies provide valuable EEG-based datasets in imagined speech paradigms, our investigation reveals that they all rely on visual or auditory cues during the data collection procedure. This setup directs the subject to simulate the target (word or command) corresponding to the presented visual or auditory cue in each trial.
However, this method contrasts with real-world scenarios, where a subject would naturally simulate a command of their choice without any additional visual or auditory cues, and these cues may affect the validity of pure imagined speech based subject identification studies.

To address these gaps, our study introduces an imagined speech paradigm for subject identification that eliminates the use of visual or auditory cues. In each trial, subjects naturally select and imagine the pronunciation of a word from a predefined list of five words without being visually or auditorily presented with the words during the trial. This approach ensures that the imagined speech process is more aligned with real-world scenarios.
Our dataset consists of over 4,350 trials collected from eleven subjects, spanning five sessions per subject conducted within a single day. This allows us to explore the robustness of models over time. Additionally, we adopt a valid evaluation methodology, ensuring that training and testing samples are drawn from different sessions causally, demonstrating the model's ability to generalize across temporal variations.

A comparison between our dataset and those from other EEG datasets with the imagined speech task is presented in Table \ref{tab:dataset_comparison}.

\begin{table}[h!]
    \centering
    \caption{Comparison of different EEG-based imagined speech datasets based on being cueless and using semantically meaningful words.}
    \label{tab:dataset_comparison}
    \resizebox{\columnwidth}{!}{%
    \begin{tabular}{|l|c|c|}
    \hline
    \textbf{Dataset} & \textbf{Cueless} & \textbf{Semantic words} \\ \hline
    
    Zhao et al. (KARA ONE) \cite{zhao_karaone} & - & - \\ \hline
    
    Brigham et al. \cite{brigham2010subject} & + & - \\ \hline

    Moctezuma et al. \cite{moctezuma2019subjects} & - & + \\ \hline

    Coretto et al. \cite{coretto2017_dataset} & - & + \\ \hline

    Asghari Bejestani et al. \cite{asghari2022_dataset} & - & + \\ \hline

    Qureshi et al. \cite{qureshi2017_dataset} & - & + \\ \hline

    BCI Competition 2020 Track3 \cite{bci_competition_dataset} & - & + \\ \hline

    \textbf{Our Study} & + & + \\ \hline

    \end{tabular}}
\end{table}


\section{METHODS AND MATERIALS}
\subsection{PARTICIPANTS}
Eleven healthy participants, all right-handed native Persian speakers with no history of neurological issues, voluntarily participated in this experiment and provided written informed consent. The group consists of 7 males and 4 females, aged between 21 and 28 years, with a mean age of 23.82 years (std $=\pm 2.44$).
In this study, participants are identified by IDs ranging from “Sub-01” to “Sub-11”. Details about each of them are provided in Table \ref{tab:participants_info}.

\begin{table*}[h!]
    \centering
    \caption{Participants information.}
    \label{tab:participants_info}
    \begin{tabular}{|l|l|l|l|l|l|l|l|l|l|l|l|}
    \hline
    \textbf{Subject ID} &
    Sub-01 & Sub-02 & Sub-03 & Sub-04 & Sub-05 & Sub-06 & Sub-07 &
    Sub-08 & Sub-09 & Sub-10 & Sub-11
    \\ \hline
    
    \textbf{Gender} &
    Male & Male & Male & Male & Male & Male & Male &
    Female & Female & Female & Female
    \\ \hline
    
    \textbf{Age} &
    23 & 24 & 22 & 24 & 21 & 21 & 24 &
    27 & 28 & 21 & 27
    \\ \hline
    \end{tabular}
\end{table*}

\subsection{EXPERIMENTAL PROTOCOL}
During the recording sessions, subjects sat comfortably in an armchair about 60 centimeters from an LCD, which was only used to signal when they should imagine the pronunciation, in a dark, silent room. To capture intra-subject variance, data was collected in five sessions on the same day for each subject. This approach can help researchers design identification systems that are more robust to intra-subject distribution shifts. The necessary instructions were provided during the installation of the EEG cap and electrodes on the subject's head.

Since the subjects are native Persian speakers, five Persian words were selected as targets. These words correspond to "Left," "Right," "Forward," "Backward," and "Stop" in English. Information about the pronunciation and the corresponding English equivalent of each word is provided in Table \ref{tab:wrods_pronounciations}.

\begin{table}[h!]
    \centering
    \caption{Target words IPA (International Phonetic Alphabet) pronunciations in Persian and their corresponding English equivalents.}
    \label{tab:wrods_pronounciations}
    \resizebox{\columnwidth}{!}{%
    \begin{tabular}{|c|c|}
    \hline
    \textbf{Word Pronunciation in Persian} & \textbf{English Equivalent} \\ \hline
    \textipa{/tS{\ae}p/} & Left \\ \hline
    \textipa{/rA:st/} & Right \\ \hline
    \textipa{/dZelo/} & Forward \\ \hline
    \textipa{/{\ae}G{\ae}b/} & Backward \\ \hline
    \textipa{/ist/} & Stop \\ \hline
    \end{tabular}
    }
\end{table}

\begin{figure*}[ht]
    \centering
    \includegraphics[width=\textwidth]{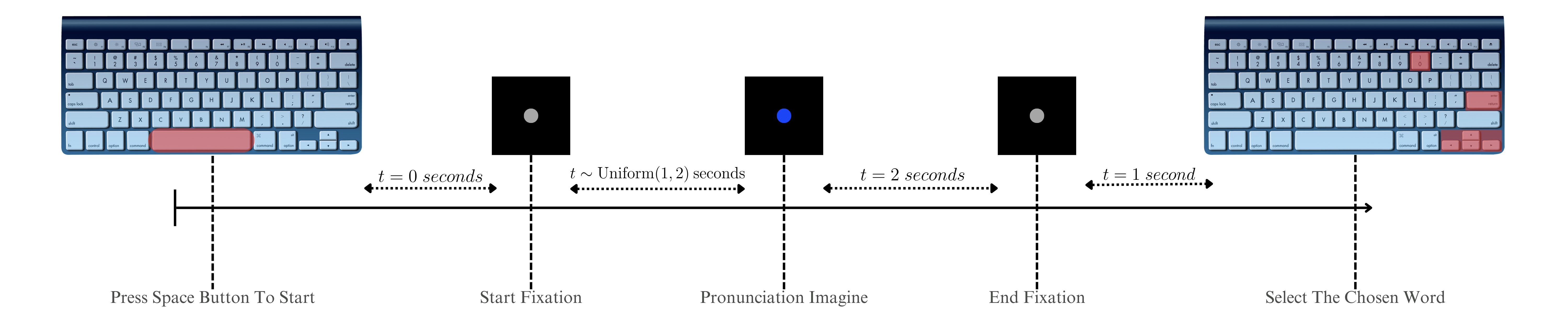}
    \caption{The experimental paradigm for a single trial. Timings specify the duration of each phase, and highlighted keys indicate permissible user inputs at different stages.}
    \label{fig:Trial_paradigm}
\end{figure*}

The experiment paradigm was developed using PsychoPy\textregistered{} \cite{psychopy_ref} software. At the start of each session, a welcome message displaying the list of words is shown on the screen for 10 seconds. We emphasize that this message is shown only at the beginning of each session, not before each trial. The subject independently selects one of the words from the list before pressing the "space" button on the keyboard to begin each trial. Since movements can affect EEG signals, fixation intervals are used to minimize this. 
After pressing the "space" button, the subject must avoid moving any body parts, including hands. A gray circle will then appear on the screen for a random duration, uniformly distributed between 1 and 2 seconds. This gap allows the subject to stabilize, ensuring that any hand movement artifacts from pressing the keyboard do not interfere with the EEG signals during the imagined speech phase in the upcoming part. This method helps minimize noise in the EEG data, making the recording more accurate.
Afterward, the circle turns blue, indicating that the subject should imagine the pronunciation of the chosen word. The subject imagines the pronunciation once, keeping their eyes open while looking at a blank black screen. The imagination begins as soon as the subject notices the color change to blue, and the blue circle remains on the screen for 2 seconds.
Next, the circle turns gray again and remains for 1 second before disappearing. The subject then sees a message on the screen instructing them to select the imagined word by pressing a specific key on the keyboard: 0 key for "Stop", the left arrow key for "Left", the right arrow key for "Right", the up arrow key for "Forward", and the down arrow key for "Backward". If any issues occurred during the 2 second imagination phase, such as unintended movements, imagining multiple words, or an incorrect imagined pronunciation, subject could press the return/enter key to mark the trial as a bad trial for removal in future processing.
The trial paradigm, along with the corresponding timings, is also illustrated in Fig. \ref{fig:Trial_paradigm}.

Through several trials and errors in designing the paradigm, the importance of the one-second end fixation became evident. Without it, subjects often began moving their hand to press the key corresponding to the imagined word before the 2-second imagination phase had finished. This early movement introduced artifacts that disturbed the EEG signal during the imagined speech phase. The one-second fixation helps prevent such movement artifacts.

The random 1 to 2 second fixation period before the imagination phase was introduced to prevent the subject from predicting the task. This variability keeps the subject focused and prepared for the upcoming task, while the gap also ensures a reduction in the likelihood of movement artifacts that could disturb the EEG signal.

Each subject participated in 5 sessions within a single day, with 100 trials in the first three sessions and 50 in the fourth and fifth sessions. The whole process takes around 5 hours for each subject.
Number of trials in the last two sessions was reduced to minimize the fatigue, which can negatively impact the quality of the EEG signals. The non-fixed trial times allow subjects to take short breaks when needed, helping to prevent fatigue during each session and maintain the quality of their EEG data. Additionally, rest intervals were provided between sessions, during which subjects could drink water and the room lights were turned on to reduce eye strain.

\subsection{DATA ACQUISITION}
\begin{figure}[ht]
\centering
\includegraphics[width=.5\linewidth]{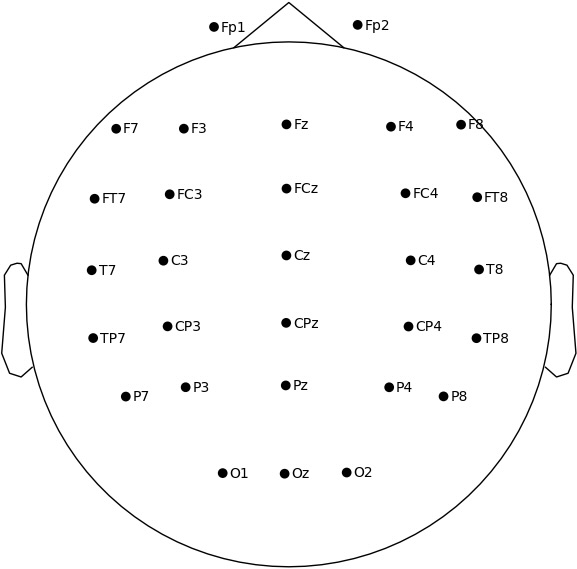}  
\caption{EEG channels used in the experiment, arranged according to the 10-20 system.}
\label{fig:montage}
\end{figure}

An active EEG device from Liv Intelligent Technology was used for data acquisition (\href{https://www.lliivv.com/en/product}{https://www.lliivv.com/en/product}). It includes EEG caps in various sizes, with electrodes positioned according to the 10-20 system \cite{10_20_system}. 
We used the device in this setup, which includes 30 EEG electrodes, one reference electrode, and two ground electrodes which are placed on the left and right earlobes. The EEG channels used in this study are illustrated in Fig. \ref{fig:montage}.The sampling rate was set to 250 Hz. For each subject, we used the appropriate size cap. To position the cap on the subject's head, the midpoint between the nasion and inion was identified, and the Cz channel was aligned with this central point.
One useful feature of this EEG device is that each EEG electrode is equipped with an LED indicator. The LED glows red when there is no proper contact with the scalp and turns blue once the electrode establishes good contact. Conductive gel was applied to each electrode to ensure proper contact and signal quality.
For each session, data from the EEG device is stored in a .tdms file, while the imagined word in each trial is stored in a .csv file.

\subsection{PRE-PROCESS}
To improve the data structure and make information extraction easier for researchers, several pre-processing steps were applied. The pre-processing was performed in Python \cite{python_ref}, primarily using the MNE library \cite{MNE_refrence}. The Python script for this part is available in the "signal\_preprocessing.py" file. Additionally, the raw device output data has been anonymized and is provided alongside the processed data, allowing researchers to adjust or modify the processing steps if needed in the future.

\subsubsection{Data structuring and file integration}
As previously mentioned, for each session, a .tdms file containing EEG data and a .csv file with the corresponding words are stored. The .tdms output file is not well-structured and includes only raw triggers indicating when the subject imagined the word. We developed a Python script to integrate the .tdms and .csv files for each session. Using MNE, we annotated the events within the data and removed trials where the subject pressed the return/enter key. Finally, the results were saved in a more organized .fif format.
The relevant code for this step can be found in the "data\_structuring\_file\_integration.py" file.

\subsubsection{Signal pre-processing}
Pre-processing includes several key techniques to enhance the signal quality, such as applying filters to remove noise, identifying and correcting bad channels, and re-referencing the data. These steps could be done to ensure the EEG recordings are properly cleaned and prepared for further analysis.

\begin{itemize}
    \item \textbf{Notch filter} \\
    Researchers commonly use Notch filter to remove power line noise, as seen in studies such as Nguyen et al. \cite{notch_ref_1} and Rusnac et al. \cite{notch_ref_2}.
    In this study, an overlap-add FIR (Finite Impulse Response) Notch filter was applied using the notch\_filter function from MNE to remove 50 Hz line noise, as well as its harmonics (e.g., 100 Hz), considering the Nyquist frequency for a sampling rate of 250 Hz.
    
    \item \textbf{Bandpass filter} \\
    To enhance the signal-to-noise ratio in this study, a FIR bandpass filter with a frequency range of 3 to 45 Hz was applied to the data. The filter utilized a Hamming window, with transition bandwidths of 2 Hz for both the low and high ends.
    
    \item \textbf{Detect and interpolate bad channels} \\
    Bad channels are those with a low signal-to-noise ratio \cite{bigdely2015prep}. This typically occurs due to poor electrode contact with the scalp.
    To keep this preprocessing pipeline fully automated and eliminate the need for manual detection of bad channels, we employed the find\_bad\_by\_correlation function from the pyprep library \cite{bigdely2015prep, appelhoff_2023_10047462}. In our setup using this function, the data is divided into non-overlapping 1-second windows, and the absolute correlation between each channel and all others is calculated for each window. A window is flagged as bad if its maximum correlation with another channel falls below the threshold of 0.4. Finally, a channel is then marked as bad if the proportion of bad windows exceeds the threshold of 0.02.
    To maintain data dimensionality across sessions and subjects, bad channels were interpolated rather than being completely removed. We used spherical splines for this interpolation, following the method by Perrin et al. \cite{perrin1989spherical}. For implementation, we utilized the interpolate\_bads function from MNE, which applies this interpolation method.
    
    \item \textbf{Re-referencing} \\
    The common average reference removes the common information recorded across electrodes at the same time, which improves the signal-to-noise ratio \cite{moctezuma2019subjects}.
    We applied this method using the set\_eeg\_reference function from MNE, with the ref\_channels parameter set to average.
    
\end{itemize}

\section{Classification Framework}
To Show the validity and evaluate our dataset and also establish comprehensive benchmarks, we investigated two distinct approaches for subject identification using pre-processed EEG signals. The first approach follows a two-stage pipeline where features are first extracted from the signals and then fed into classifiers. The second approach uses end-to-end deep learning architectures, specifically designed for EEG classification, that learn both feature representations and classification boundaries directly from the EEG signals to identify the subjects. These approaches are detailed in the following sections.

\subsection{Two-Stage Classification Approaches}
Feature extraction followed by shallow classifiers is an established approach in EEG-based subject identification. In this work, we employ two methods for feature extraction. The first method involves manual extraction of domain-specific features from the preprocessed EEG signals, utilizing statistical and signal processing techniques. The second method leverages deep feature extraction, where we fine-tune a foundational time-series deep learning model to extract embeddings from the preprocessed EEG signals. These extracted features are then fed into shallow classification algorithms, which are trained to distinguish between subjects based on the computed features.

For shallow classification, we utilized two machine learning models: Support Vector Machine (SVM) with a radial basis function (RBF) kernel \cite{cortes1995support, scholkopf2002learning} and XGBoost \cite{chen2016xgboost} to classify subjects based on the extracted features. These classifiers were chosen to evaluate the effectiveness of extracted features for subject identification in this dataset.

\subsection{Manual Feature Extraction}
We computed five categories of features from the EEG signals which are statistical-based features, wavelet-based features, mel-frequency cepstral coefficients (MFCC), power spectral density (PSD) and autoregressive (AR) coefficients. For statistical features, we extracted a set of time-domain characteristics including mean, variance, skewness, and kurtosis for each EEG channel. These features capture the statistical properties of the signal's amplitude distribution over time.
For wavelet-based features, we computed the energy of wavelet decomposition coefficients for each EEG channel. This approach captures both temporal and spectral characteristics of the signal, providing a suitable representation of the EEG dynamics.

In addition to these, we incorporated MFCC, PSD, and AR coefficients, which have been widely adopted in identity recognition studies. Prior works such as those by Huang et al. \cite{huang2022m3cv}, Arnau-Gonzalez et al. \cite{arnau2018influence}, and Nakamura et al. \cite{nakamura2017ear} have demonstrated the discriminative potential of these features for subject identification.

\subsection{Deep Feature Extraction}
For this method,
we explored the use of time-series foundation models to extract embeddings directly from EEG signals, rather than relying on handcrafted features. We utilized the MOMENT \cite{goswami2024moment} model , a transformer-based architecture, and fine-tuned it on our pre-processed dataset to derive meaningful embeddings. These embeddings are designed to capture distinctive characteristics of the EEG signals, enabling effective discrimination between subjects within the embedding space.

\subsubsection{MOMENT}
MOMENT is a transformer-based model pre-trained on a large collection of public time-series data using a masked prediction method. This self-supervised learning task involves masking portions of the input data and training the model to accurately reconstruct the missing parts. By leveraging this technique, Moment captures intrinsic patterns and characteristics of time-series data.

The authors of MOMENT pre-trained the model using three different architectures for the encoder: T5-Small, T5-Base, and T5-Large. This results in MOMENT having three distinct configurations: the Small model with an embedding dimension of 512, the Base model with an embedding dimension of 768, and the Large model with an embedding dimension of 1024.

The pre-training approach allows Moment to generalize across various types of time-series datasets. The authors demonstrated that, even without dataset-specific fine-tuning, Moment can learn distinct and class-specific representations for diverse time-series data across different fields. In our study, we fine-tuned different sizes of Moment on our dataset to adapt the embeddings for subject identification, utilizing its capability to capture distinctive features from EEG signals that are relevant for distinguishing between subjects.

\subsection{Shallow Classifiers}
For classification using the extracted features, we employed two established machine learning algorithms SVM and XGBoost. These algorithms, often referred to as shallow classifiers to distinguish them from deep learning approaches, have demonstrated strong performance in EEG classification tasks \cite{parui2019emotion, bousseta2016eeg}. SVM excels in high-dimensional feature spaces and can effectively handle non-linear relationships through kernel functions, while XGBoost provides robust performance through ensemble learning of decision trees.

SVM is a supervised learning algorithm which aims to find the optimal hyperplane that maximizes the margin between two classes in the feature space. Given the training data $\{(x_1, y_1), ..., (x_n, y_n)\}$ SVM solves the following optimization problem :

\begin{align}
    \min_{w,b} \frac{1}{2} ||w||^2 \text{ subject to } y_i(w^Tx_i - b) \geq 1, \forall i
\end{align}
where $w$ is the weight vector, $b$ is the bias term, and $y_i$ is the class label. Then the decision function to classify the data is :
\begin{align}
    f(x) = \text{sgn}(w^T x - b)
\end{align}
SVM could also use a kernel function in cases of non-linearly separable data. in this study the radial basis function (RBF) is used as the kernel which is defined as :
\begin{align}
    K(x, x') = exp (-\frac{||x - x'||^2}{2\sigma^2})
\end{align}

Our second shallow classifier, XGBoost is an efficient and scalable implementation of gradient boosting based on the work by Friedman \cite{friedman2001greedy}. XGBoost builds an ensemble of weak learners, in this study decision trees, by minimizing the following loss function :
\begin{align}
    L = \sum_{i=1}^N \ell(y_i, \hat{y}_i) + \sum_{k=1}^K \Omega(f_k)
\end{align}
Where $\ell(y_i, \hat{y}_i)$  is the loss function that measures the difference between the predicted output $\hat{y}_i$ and the true output $y_i$, $\Omega(f_k)$ is the regularization term that penalizes the complexity of the model (i.e. the complexity of each tree $f_k$), N is the number of training samples and K is the number of decision trees. Finally the prediction for an input instance is given by :
\begin{align}
    \hat{y}_i = \sum_{k=1}^K f_k(x_i), f_k \in F
\end{align}
where k is the number of trees, $f_k$ is a function in the functional space $F$, and $F$ is the set of all possible decision trees.

\subsection{End-to-End Deep Learning Approaches}
End-to-end deep learning models have become increasingly popular in EEG signal processing because they can learn optimal feature representations directly from data, eliminating the need for manual feature engineering. This automated feature learning can potentially capture subtle patterns and dependencies in EEG signals that might be overlooked by traditional handcrafted features. Additionally, by jointly optimizing feature extraction and classification, these models can adapt to dataset-specific characteristics and potentially achieve better performance.

Here, we utilized several deep learning architectures that have been specifically introduced for EEG classification and brain-computer interface (BCI) tasks. The models employed include EEGNet \cite{lawhern2018eegnet}, Shallow ConvNet \cite{schirrmeister2017deep} and EEG conformer \cite{song2022eeg}. The architecture of these models is designed to automatically capture complex discriminative features of EEG signals with respect to the target classes. In our study we applied these models directly to the pre-processed EEG signals to identify subjects.

\subsubsection{EEGNet}
EEGNet is a compact convolutional neural network (CNN) specifically designed for EEG-based BCIs \cite{lawhern2018eegnet}. It utilizes separable convolutions, which reduce the number of trainable parameters, making the model more compact without decreasing the performance. EEGNet leverages spatial and temporal convolutions that are applied to EEG signals, which have distinctive frequency and spatial characteristics.

The architecture is structured in two main blocks. Block 1  applies temporal convolutions followed by depthwise spatial convolutions to capture both temporal dynamics and spatial relationships across EEG channels. Block 2 employs separable convolutions to further decrease model complexity while combining the extracted features. The output features are then passed through a softmax classifier to perform the final classification.

\subsubsection{Shallow ConvNet}
Shallow ConvNet is a CNN architecture inspired by the Filter Bank Common Spatial Patterns (FBCSP) approach introduced by Ang et al. \cite{ang2008filter}. Unlike FBCSP, which separates feature extraction and classification into distinct stages, Shallow ConvNet integrates these stages within a single network, allowing the model to learn all transformations jointly. The architecture begins with temporal convolution and spatial filtering layers, similar to the bandpass and common spatial pattern (CSP) steps in FBCSP. Following these transformations, the model applies a squaring nonlinearity, mean pooling, and a logarithmic activation function, which together replicate the log-variance computation performed in the FBCSP pipeline.
Additionally the Shallow ConvNet use of multiple pooling regions within a trial allows it to learn the temporal structure of band power changes, which enhances classification performance.

\subsubsection{EEG Conformer}
EEG Conformer is a novel convolutional transformer architecture which combines convolutional layers with self-attention mechanisms \cite{vaswani2017attention} to capture both local and global features of EEG signals. Unlike conventional convolutional networks that typically focus on local temporal patterns, EEG Conformer incorporates a self-attention module that enables the model to capture long-range dependencies. The convolution module is responsible for learning low-level temporal and spatial features, while the self-attention module extracts global correlations from these local features, allowing the model to understand complex interdependencies in EEG signals.

\section{Results}
In this section,
we present the results of applying the above methods to the pre-processed EEG signals for the subject identification task. In all approaches, each subject is treated as a separate class. To prevent data leakage and ensure proper training, validation, and testing, we adopted a session-based hold-out validation strategy. Specifically, we used the first three sessions for training, the fourth session for validation, and the final session for testing. this procedure ensures better robustness over time variations and improves the model's generalization ability.
The results are detailed in the subsequent sections.

\subsection{Identification Based On Feature Extraction}
The performance of shallow classifieres, including SVM with an RBF kernel and XGBoost, is evaluated using a variety of feature extraction methods applied to the preprocessed EEG signals. These feature sets included statistical-based features, wavelet-based features, MFCC, PSD, AR coefficients, and embeddings extracted from the MOMENT model.
The statistical and wavelet-based features were extracted using the mne-features library \cite{schiratti2018ensemble}.
MFCC features were computed using the librosa \cite{mcfee2015librosa} library by applying short-time analysis to each EEG channel and averaging cepstral coefficients across frames. PSD features were derived using Welch’s \cite{welch1967use} method, with power spectra interpolated into a fixed-length vector of 30 frequency bins per channel. AR features were extracted by fitting an autoregressive model to each channel using the statsmodels \cite{seabold2010statsmodels} library.

For the implementation of the SVM we used scikit-learn library \cite{scikit-learn} which in the case of multiclass classification, the library handles it using a one-vs-one scheme. In this scheme, a binary classifier is trained for each pair of classes, and the class predicted by the majority of classifiers is chosen as the final classification.
The shallow classifiers have hyperparameters that control their complexity and directly affect their performance. To optimize these hyperparameters, the Optuna \cite{optuna_2019} framework was utilized.

To perform subject identification using embeddings extracted by MOMENT, we fully fine-tuned the MOMENT model, including its classifier head, to better adapt it to the task and extract more discriminative embeddings. This fine-tuning process aimed to optimize the model's parameters so that the embeddings could capture features that are more relevant for distinguishing between subjects.
For fine-tuning, we followed the hyperparameters suggested by the authors of the MOMENT model. The convergence of the fine-tuning process was observed after 35, 32, and 51 epochs for the large, base, and small configurations, respectively. To evaluate the performance of shallow classifiers on the embeddings extracted by MOMENT, we also removed the classifier head after fine-tuning. This allowed us to feed the extracted embeddings directly into shallow classifiers to assess their effectiveness in subject identification.
The results for shallow models applied on different feature sets, are presented in Table \ref{tab:base_models}.

The results in Table \ref{tab:base_models} highlight several key findings. The combination of wavelet-based features with an RBF SVM classifier achieved the highest accuracy at 94.17\%, significantly outperforming other methods in this category. This suggests that time-frequency characteristics captured by wavelets are highly discriminative for this subject identification task.

Regarding the performance of different-sized MOMENT models, a counter-intuitive trend emerges. For both the RBF SVM and the head classifier, performance shows an inverse relationship with model size. The MOMENT-small model yields the best results, followed by base, and then large. A similar pattern is observed with XGBoost, where the large model is again the least effective. This suggests that larger models and embedding sizes of the MOMENT, despite their increased capacity, may be more prone to overfitting during the fine-tuning stage, leading to poorer generalization on the unseen test session.

\begin{figure}[ht]
\centering
\includegraphics[width=\linewidth]{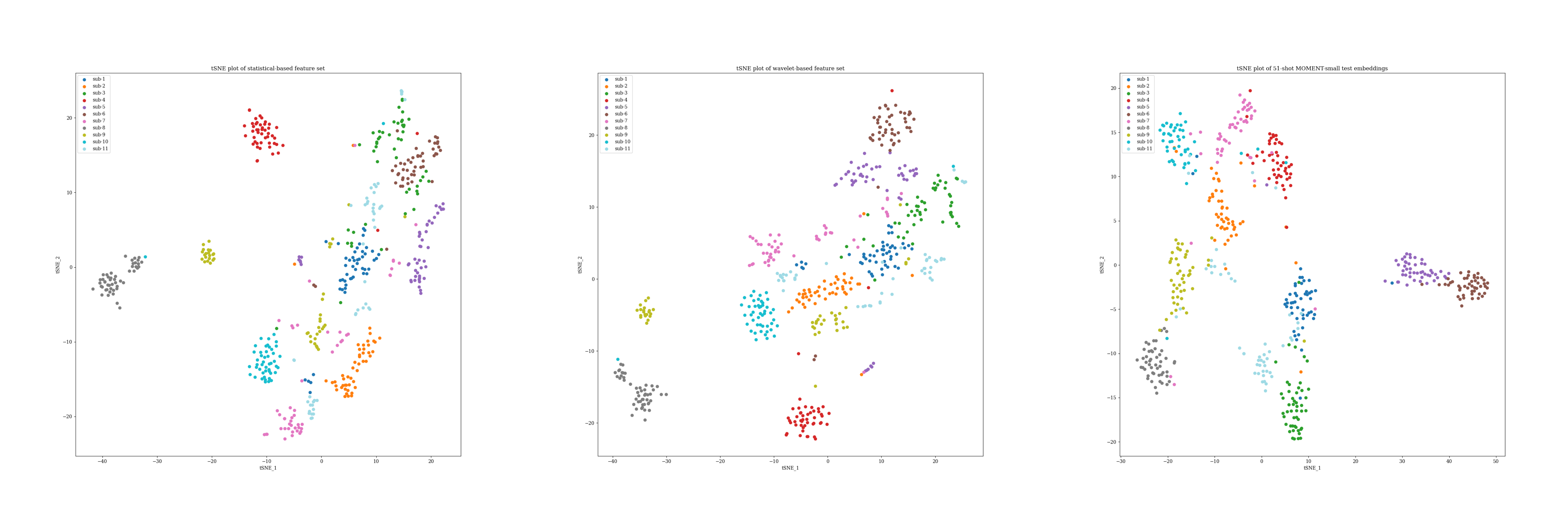}
\caption{
t-SNE visualizations of the extracted features in a 2-dimensional space for subject identification. From left to right, the subplots illustrate statistical features, wavelet features, and fine-tuned MOMENT-small embeddings, respectively.
}
\label{fig:tsne_features}
\end{figure}

\begin{table*}[h!]
    \centering
    \caption{Performance of subject identification based on applying shallow classifiers to manual extracted features.}
    \label{tab:base_models}
    \begin{tabular}{|c|c|c|c|c|}
    \hline
    
    \textbf{Model} & \textbf{Model input} & \textbf{Accuracy(\%)} & \textbf{Precision(\%)} & \textbf{Recall(\%)} \\ \hline

    \multirow{8}{*}{RBF SVM} 
    & statistical-based features & 84.40 & 85.34  & 84.45  \\ \cline{2-5}
    
    & wavelet-based features& \textbf{94.17}  & \textbf{94.64}  & \textbf{94.16}  \\ \cline{2-5}

    & MFCC & 89.85 & 91.72   & 89.82   \\ \cline{2-5}

    & PSD & 85.15 & 91.43   & 85.07  \\ \cline{2-5}

    & AR coefficients &  80.83 & 82.09   & 80.80  \\ \cline{2-5}

    & MOMENT-small embeddings & 89.47  & 89.78  & 89.51  \\ \cline{2-5}

    & MOMENT-base embeddings & 88.72   & 88.72 & 88.74    \\ \cline{2-5}
    
    & MOMENT-large embeddings & 87.41 & 87.70  & 87.42  \\ \hline

    \multirow{8}{*}{XGBoost} 
    & statistical-based features& 77.82  & 82.16  & 77.94  \\ \cline{2-5}
    
    & wavelet-based features&  80.45  & 79.93  & 80.47  \\ \cline{2-5}

    & MFCC & 82.52 & 82.16   & 82.48   \\ \cline{2-5}

    & PSD & 87.41 & 87.74   & 87.48  \\ \cline{2-5}

    & AR coefficients &  79.70 & 81.38   & 79.69  \\ \cline{2-5}

    & MOMENT-small embeddings & 86.09  & 86.81  & 86.12  \\ \cline{2-5}

    & MOMENT-base embeddings & 87.22    & 87.30    & 87.21    \\ \cline{2-5}

    & MOMENT-large embeddings & 82.71  & 83.18  & 82.76  \\ \hline
    
    \multirow{3}{*}{head classifier} 
    & MOMENT-small embeddings & 88.53  & 88.90  & 88.56  \\ \cline{2-5}
    
    & MOMENT-base embeddings & 87.78  & 87.77  & 87.80  \\ \cline{2-5}

    & MOMENT-large embeddings & 86.65  & 86.76  & 86.66  \\ \hline

    \end{tabular}
\end{table*}

t-SNE \cite{van2008visualizing} is also used to plot a subset of extracted feature sets of the test data in a 2-dimensional space in Fig. \ref{fig:tsne_features}. The plot visually demonstrates how different feature sets are distributed for different subjects in the test data, providing insight into the separability and clustering of the data in a lower-dimensional space.

An apparent discrepancy exists between the t-SNE visualizations in Fig. \ref{fig:tsne_features} and the classification results in Table \ref{tab:base_models}. MOMENT embeddings form cleaner 2D clusters, yet wavelet-based features yield better classification performance. This is because t-SNE reduces data to two dimensions, often distorting the structure relevant for classification, whereas the SVM operates in the original high-dimensional space, capturing richer discriminative patterns. Thus, while MOMENT features are visually well-structured in 2D, wavelet features retain more useful information for accurate classification.

\subsection{End-to-End Subject Identification}
In this section, we trained three deep learning architectures EEGNet, Shallow ConvNet, and EEG Conformer directly on the pre-processed EEG signals, which were described in the previous section. To implement these models effectively, we utilized the Braindecode library \cite{HBM:HBM23730}, which provides optimized and efficient implementations of these deep learning architectures in Python.

For the training process, the hyperparameters recommended by the Braindecode library were adopted for each model. However, the learning rate, weight decay, as well as training batch size, were optimized using the Optuna framework. The validation dataset was employed for this hyperparameter tuning. Furthermore, training was capped at a maximum of 30 epochs, with early stopping implemented based on the validation accuracy. Specifically, training was stopped if there was no improvement in validation accuracy for six consecutive epochs.
The results of these models on the test dataset are presented in Table~\ref{tab:deep_models}.

\begin{table*}[h!]
    \centering
    \caption{Subject identification results using different end-to-end deep learning models.}
    \label{tab:deep_models}
    \begin{tabular}{|c|c|c|c|c|}
    \hline
    
    \textbf{Model} & \textbf{Model Input} & \textbf{Accuracy(\%)} & \textbf{Precision(\%)} & \textbf{Recall(\%)} \\ \hline
    
    EEGNet & pre-processed EEG &  84.77  & 86.95  & 84.89  \\ \hline

    Shallow ConvNet & pre-processed EEG & \textbf{99.44}  & \textbf{99.45}  & \textbf{99.43}  \\ \hline

    EEG Conformer & pre-processed EEG & 97.37  & 97.52  & 97.39  \\ \hline 
    
    \end{tabular}
\end{table*}

As detailed in Table~\ref{tab:deep_models}, the Shallow ConvNet architecture stands out by achieving the highest accuracy of any method in this study, reaching a remarkable $98.87\% \pm 0.41\%$. It is also noteworthy that the standard deviation for all end-to-end models in this table is less than 1\%.

\subsection{Further Analysis}
Based on the high performance of the end-to-end models and the strong results from the feature-based SVM, we selected the top three models which are Shallow ConvNet, EEG Conformer, and SVM with wavelet features for a more in-depth evaluation. To better understand their practical capabilities and robustness, we conducted a series of further analyses which will be presented in the following sections.

\subsubsection{Subject-Specific Performance Analysis}
To evaluate the robustness of the models across individuals, we computed per-subject accuracies for the top models used in this study. The results are presented in Table~\ref{tab:per_subject_acc}.

This analysis reveals that the Shallow ConvNet model demonstrates consistently high performance across all subjects and exhibiting the lowest variance among the three models.

By contrast, the SVM model shows greater fluctuation in accuracy, particularly for Subject 6, where performance drops significantly. EEG Conformer performs well overall but displays slightly more variation than the Shallow ConvNet.

\begin{table*}[h!]
    \centering
    \caption{Per-subject Accuracy (\%) for Top Models.}
    \label{tab:per_subject_acc}
    \resizebox{\textwidth}{!}{%
    \begin{tabular}{|c|c|c|c|c|c|c|c|c|c|c|c|}
    \hline
    \diagbox{\textbf{Model}}{\textbf{Subject}} & Sub-01 & Sub-02 & Sub-03 & Sub-04 & Sub-05 & Sub-06 & Sub-07 & Sub-08 & Sub-09 & Sub-10 & Sub-11 \\ \hline
    SVM + wavelet features & 100.0 & 97.92 & 95.92 & 97.92 & 87.50 & 72.92 & 94.0 & 97.96 & 95.74 & 95.83 & 100.0 \\ \hline
    Shallow ConvNet & \textbf{100.0} & \textbf{100.0} & \textbf{100.0} & \textbf{100.0} &\textbf{95.83} & \textbf{100.0} & \textbf{100.0} & \textbf{100.0} & \textbf{100.0} & \textbf{97.92} & \textbf{100.0} \\ \hline
    EEG Conformer & 100.0 & 97.92 & 89.80 & 97.92 & 95.83 & 100.0 & 98.00 & 100.0 & 100.0 & 97.92 & 93.88 \\ \hline
    \end{tabular}
    }
\end{table*}

\subsubsection{Impact of Target Words on Subject Identification}
To explore how different imagined words contribute to subject identification performance, we evaluated the accuracy of top models separately for each target word. The results of this word-level analysis are shown in Table~\ref{tab:per_word_acc}.

The analysis reveals that the imagined word \textit{“stop”} (corresponding to the Persian word pronounced /ist/) consistently achieved the highest accuracy across all three models. This suggests that \textit{“stop”} is the most discriminative target word for subject identification in our dataset.

\begin{table*}[h!]
    \centering
    \caption{Accuracy Of The Subject Identification Per Target Word For Top Models}
    \label{tab:per_word_acc}
    \begin{tabular}{|c|c|c|c|}
    \hline
    
    \textbf{Target Word (in English)} & \textbf{SVM + wavelet features} & \textbf{Shallow ConvNet} & \textbf{EEG Conformer} \\ \hline

    Backward & $91.26$ &  $100$ & $98.06$ \\ \hline

    Forward & $94.55$ &  $100$ & $98.18$ \\ \hline

    Left & $94.17$ &  $98.06$ & $98.06$  \\ \hline
    
    Right & $94.83$ &  $99.14$  & $93.97$ \\ \hline

    Stop & \textbf{96.0} & \textbf{100}  & \textbf{99.0}  \\ \hline
    
    \end{tabular}
\end{table*}

\subsubsection{Model Scalability Analysis}
To evaluate the models scalability, we examined how their performance changes as the number of subjects increases.
We conducted a series of experiments in which the number of subjects in the dataset was incrementally increased from 2 to 11. For each configuration, models were trained on data from the first three sessions, validated on the fourth session, and tested on the fifth.

\begin{figure}[ht]
\centering
\includegraphics[width=1.0\linewidth]{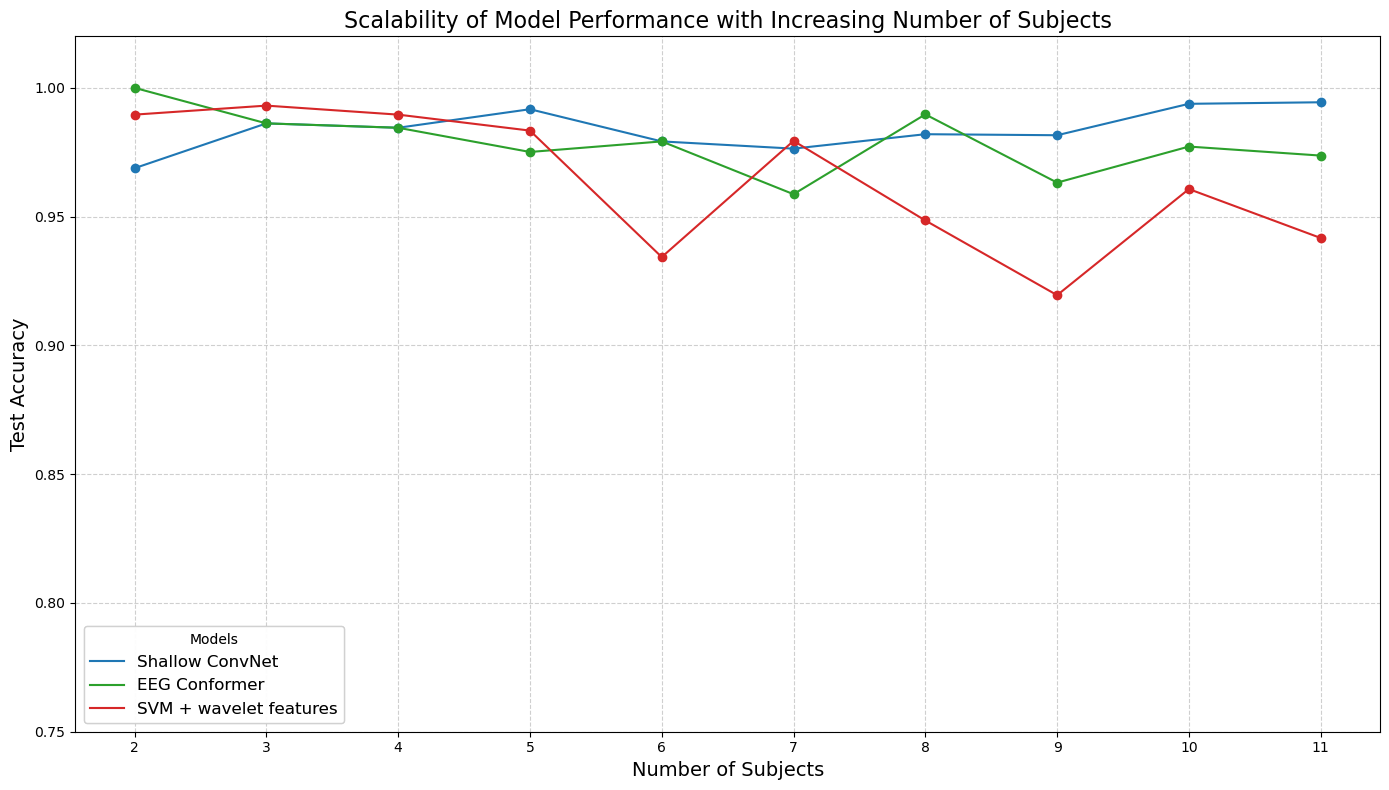}  
\caption{Test accuracy of Shallow ConvNet, EEG Conformer, and SVM with wavelet features as the number of subjects in the dataset increases from 2 to 11. Shallow ConvNet maintains high and consistent performance, EEG Conformer remains competitive, while SVM + wavelet features shows a more significant drop in accuracy, indicating limited scalability.}

\label{fig:scalability}
\end{figure}

The results of this analysis are visualized in Figure~\ref{fig:scalability}. Among the evaluated models, Shallow ConvNet demonstrated remarkable stability and scalability, maintaining test accuracies almost above 97\% across all subject counts.

\subsubsection{Effect of Temporal Gap Between Training and Testing Sessions}
To investigate the impact of temporal distance between training and testing sessions, we designed an experiment that isolates the effect of time gap while keeping the amount of training data fixed. This helps assess how temporal factors influence model generalization.

Models are trained separately on data from Session 1, Session 2, and Session 3. These sessions contain approximately the same number of samples (1032, 1064, and 1059), ensuring a fair comparison in terms of data volume.

\begin{figure}[ht]
\centering
\includegraphics[width=1.0\linewidth]{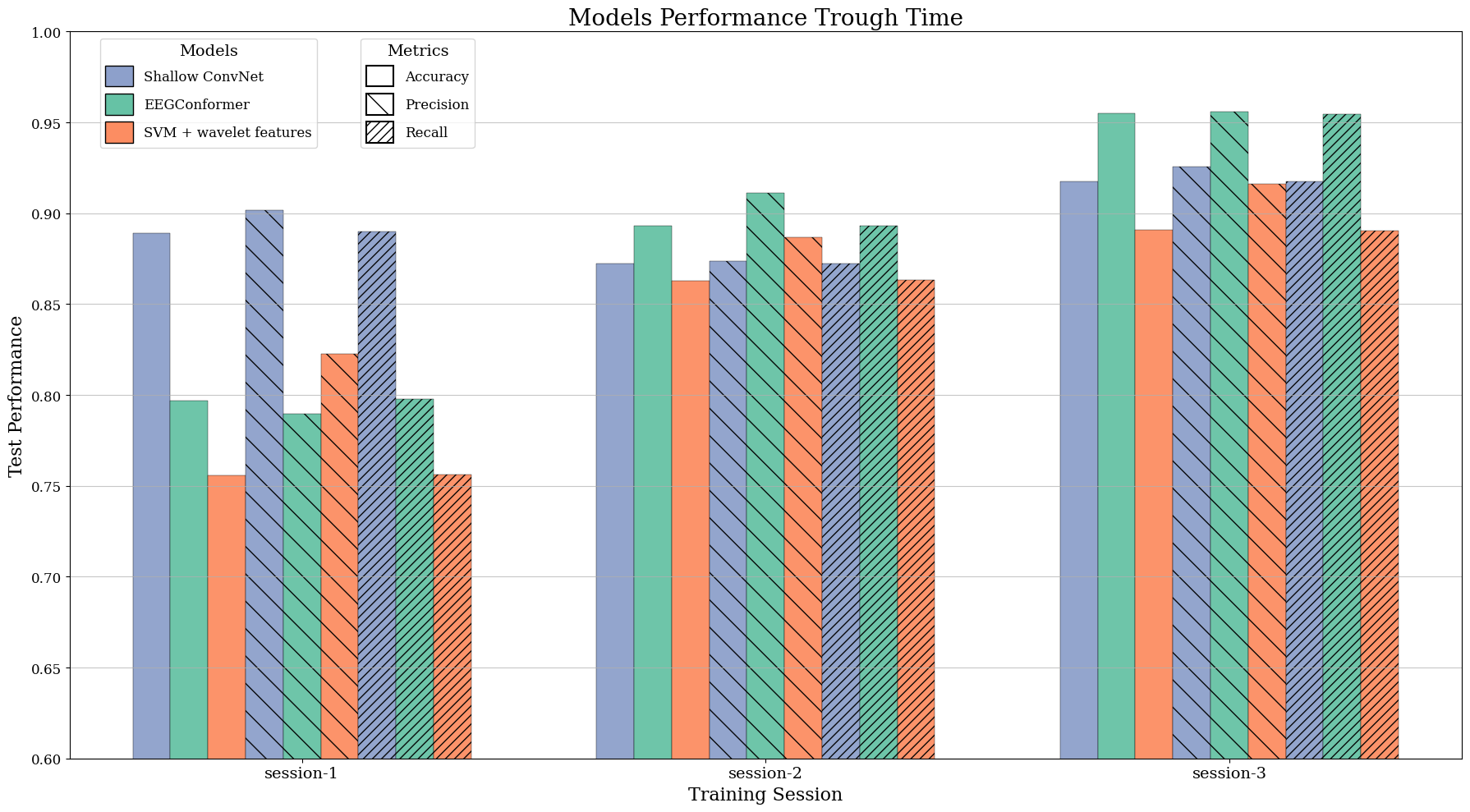}  
\caption{Comparison of the time gap between train and test. From the left to right the gap between the training and the test session is decreasing and also the performance of the models are increasing. Here the amount of the data is almost the same between sessions. The Shallow ConvNet shows to be the most robust among other end-to-end models.}
\label{fig:time_effect}
\end{figure}

Figure~\ref{fig:time_effect} illustrates that as the time gap between training and testing increases, model performance declines, highlighting the impact of temporal drift in EEG signals. Among the models, Shallow ConvNet shows the most robustness over time, with relatively stable performance compared to EEG Conformer and SVM, which experience more significant degradation.

\section{Limitations and Future Work}
Although this study presents promising results, several limitations should be acknowledged.

First, the dataset includes 11 subjects, which limits the generalizability of the findings to larger populations. While the models demonstrated strong performance and robustness within this sample, future work should expand the dataset to a more diverse and extensive participant pool to better assess scalability and generalization.

Second, the EEG data were collected using a laboratory-grade active EEG system, which is not suitable for mobile applications. Future research could explore the use of portable EEG systems, including emerging in-ear EEG technologies \cite{nakamura2017ear}, which may offer more practical solutions for subject identification in naturalistic settings.

Third, although our dataset includes multiple sessions per subject, all sessions were recorded within a single day. This limits the assessment of long-term stability. In future work, we plan to collect EEG data across multiple days or weeks to evaluate temporal robustness and better simulate real-world deployment conditions.

\section{Conclusion}
In this study, we introduced a novel cueless EEG paradigm for subject identification, where participants imagine words without external cues. Based on this paradigm, we created a comprehensive dataset containing over 4,350 trials from 11 subjects across multiple sessions. To validate the dataset and establish performance benchmarks, we conducted classification experiments using a broad spectrum of classification approaches, ranging from traditional feature-based methods to state-of-the-art deep learning architectures.
Our evaluation methodology employed session-based hold-out validation, where models were trained and tested on data from different recording sessions. This approach ensures the practical applicability of our results by demonstrating the models ability to generalize across temporal variations. The experimental results showed classification accuracies up to 99.44\% and strong model robustness, validating our dataset quality and the effectiveness of modern EEG-based subject identification methods.

\section{Code \& data Availability}

The dataset utilized in this study is publicly accessible at \url{https://huggingface.co/datasets/Alidr79/cueless_EEG_subject_identification}.All scripts and codes are available in the GitHub repository at \url{https://github.com/Alidr79/cueless_EEG_subject_identification}.

\section{Conflict of Interest}
None of the authors have a conflict of interest to disclose.


%





\ifCLASSOPTIONcaptionsoff
  \newpage
\fi





\bibliographystyle{IEEEtran}
\bibliography{IEEEabrv,Bibliography}

\end{document}